\documentclass{article}
\usepackage{spconf,amsmath,graphicx}
\usepackage[table,xcdraw]{xcolor}
\usepackage{booktabs} 
\usepackage{graphicx}
\usepackage{hyperref}
\usepackage[font=footnotesize]{subfig}

\graphicspath{ {./pictures/} }
\newcommand{\comment}[1]{}
\usepackage{amssymb}
\usepackage{xcolor}
\usepackage{hyperref}
\usepackage{booktabs}

\usepackage{xcolor}
\pagecolor{white}

\title{Tiny-CRNN: Streaming Wakeword Detection in a Low Footprint Setting}
\name{\begin{tabular}{c}Mohammad Omar Khursheed $^{1,2}$, Christin Jose$^{2}$, Rajath Kumar$^{2}$\\ Gengshen Fu$^{2}$, Brian Kulis$^{2,3}$, Santosh Kumar Cheekatmalla$^{2}$\end{tabular}}
\begin{document}
%
\maketitle
\begin{abstract}
In this work, we propose Tiny-CRNN (Tiny Convolutional Recurrent Neural Network) models applied to the problem of wakeword detection, and augment them with scaled dot product attention. We find that, compared to Convolutional Neural Network models, False Accepts in a 250k parameter budget can be reduced by 25\% with a 10\% reduction in parameter size by using models based on the Tiny-CRNN architecture, and we can get up to 32\% reduction in False Accepts at a 50k parameter budget with 75\% reduction in parameter size compared to word-level Dense Neural Network models. We discuss solutions to the challenging problem of performing inference on streaming audio with this architecture, as well as differences in start-end index errors and latency in comparison to CNN, DNN, and DNN-HMM models.
\end{abstract}
\section{Introduction}

Keyword spotting, or in the context of voice assistants such as Alexa,  \emph{wakeword detection}, is the act of detecting the presence of a certain phrase in a stream of audio. This is an important problem with applications in a variety of acoustic environments, varying from dedicated devices such as Amazon's Echo to being embedded within third-party devices such as headphones. A common architectural choice for wakeword detection is one with two stages. The first stage uses small, efficient wakeword detection models on edge devices, whose detections are then sent to the cloud, where a larger model, unconstrained by the memory and compute restrictions of edge devices, performs wakeword verification. While there has been previous work on using CRNNs for the verification task \cite{rajathpaper}, our work focuses on wakeword detection models for use on-device, which is often a much more constrained use-case. The key problem to be solved by wakeword detection models is that of minimizing the number of False Accepts (for example, a voice assistant is activated when it is not intended to be) by the device while still not being too restrictive and allowing valid wakewords. The former is important because devices, upon detecting a wakeword, usually provide the user with visual or auditory feedback, such as an Echo's light ring which turns blue while sending data to the cloud. Inadvertent activation of these systems is detrimental to the trust that customers put into such systems. The latter is important from a customer experience perspective. \\ 
It is important to note that, for models that run on edge devices, the model footprint must be small enough to fit in the memory and compute capacities available. The number of trainable \emph{parameters} that a model has, and how many \emph{multiplication operations}, or \emph{multiplies} it takes to do inference on a single piece of audio are metrics used to measure model footprint. The former is what constitutes the amount of memory the model will take when it is deployed to a device, while the latter tells us how powerful the compute systems on the device need to be. With wakeword detection usually being coupled with other tasks and very rarely being the only function of the device; these factors are of paramount importance. The more compact a model, in terms of both memory and compute, the smaller the device it can fit on in terms of these metrics. \\
Originally, wakeword detection was done through large vocabulary ASR systems \cite{115555}, which require HMMs to model entire lexicons. There has since been work using 2-stage DNN-HMM systems for wakeword detection \cite{266505}, which has been furthered through improved training strategies \cite{Panchapagesan2016MultiTaskLA}. More recently, CNN architectures have found their way into the literature, \cite{43969}, where their efficacy has been proved over traditional 2-stage DNN-HMM models, and they have found use in on-device scenarios via ideas such as depthwise convolutions \cite{885807b48e39422fa3b82f00613e0a99}. Recurrent neural networks (RNNs), a class of neural networks used to process sequential data, have also been found to be useful for keyword spotting tasks \cite{770310}, which is furthered through the use of recurrent neural network transducer (RNN-T) models, which learn both acoustic and language model components \cite{He2017StreamingSK} to improve performance in wide-ranging acoustic environments. \\
\begin{figure}[!htpb]
\centering
	\includegraphics[width=8cm,height=4cm]{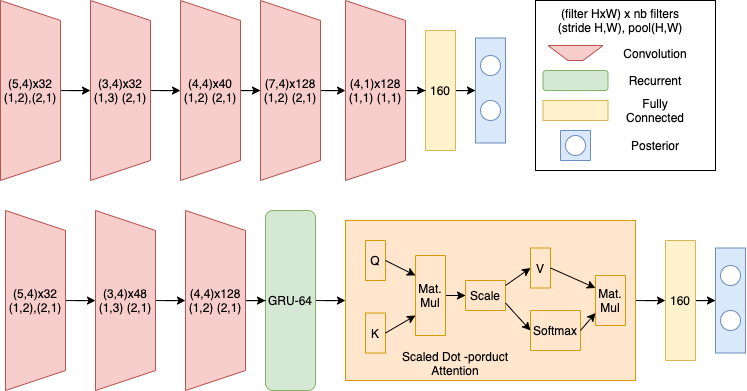}	
\caption{\emph{Top: CNN-263k baseline architecture, Bottom: Tiny CRNN architecture}}
\label{fig:archs}
\end{figure}
\\ The class of models we consider in this work take the benefits of both CNNs (of learning local features at different scales), as well as those of RNNs (learning transitions between different parts of a sequence), and combine the two types of architectures to form Convolutional Recurrent Neural Networks (CRNNs). There has previously been work using CRNNs for wakeword detection in a compute-constrained environment  \cite{DBLP:journals/corr/ArikKCHGFPC17} \cite{wei2021edgecrnn}. However, in this work, apart from building CRNN architectures at significantly lower parameter sizes \cite{wei2021edgecrnn} or outperform previous CRNN architectures \cite{DBLP:journals/corr/ArikKCHGFPC17} as well as baseline CNN and DNN models at the 250k budget level, we motivate and justify the use of attention-augmented CRNNs, which we refer to as the Tiny-CRNN architecture. We develop CRNN models based on this architecture which are robust to AFE (Audio Front End) gain changes, which are important for the wide variety of acoustic environments of the devices these models run on. We also go further, training CRNNs at a 50k parameter budget, while previous work has focused on models with a larger footprint. We follow this up with a discussion on how the convolutional front end's receptive field affects performance. We show how streaming inference can be done correctly for CRNNs through the use of parallel implementation of recurrent layers and discuss how attention makes a difference to performance with CRNNs. Overall, we find that CRNNs are effective alternatives to pure CNNs, and that a case exists for them to replace their fully convolutional counterparts in production applications. 
\section{System Design}
\label{sec:format}
\subsection{ Input Frames and Baseline Models}
The 99th percentile length of the wakeword ``Alexa" in our dataset is around 1 second. We, therefore, use 100 input frames for all our models, computing Log Mel Filter Bank Energies (LFBEs) every 10 ms over a window of 25 ms, which translates to approximately 1 second of audio. We did not attempt to use a bigger context, since that would cost us in terms of multiplies, and add negligible information in terms of improving wakeword detection abilities. l. For the 250k parameter budget, our baselines use a CNN architecture (called CNN-263k, see Figure \ref{fig:archs}, due to the 263k parameters it has; we adopt similar naming conventions for our models, except for that which refers to previous work, where we include the author name as part of the model name) and has 5 convolutional layers followed by one fully connected layer, and another similar CNN (CNN-197k) with 4 convolutional layers. For the 50k budget, we use DNNs with 6 fully connected layers, (DNN-233k and DNN-51k) and a CNN (CNN-28k) with 5 convolutional layers and 1 fully connected layer as baselines.

\subsection{Tiny-CRNN Architecture}
\label{crnn_models}
We build on the architecture in \cite{rajathpaper}, with the convolutional front-end taking the 100 frame input frames $I \in \mathbb{R}^{t \times f \times 1}$, where $t$ and $f$ are time and frequency respectively,and outputting embeddings  $D \in \mathbb{R}^{t' \times f' \times c'}$ where $c'$ is the number of output channels of the last convolutional layer, and flatten the last two dimensions, $f'$ and $c'$, to send a temporally preserved input of $D' \in \mathbb{R}^{t' \times f'c'}$ to the recurrent layer with $d$ cells. This gives us an output $L \in \mathbb{R}^{t' \times d}$, which we then pass through scaled dot product attention, which weights how important each specific timestep is. The attention block processes this input through three linear layers, the outputs of which are the key $K$ ,  the query $Q$, and  the value $V$. The dimensions of these linear layers are $d_K=d_Q=d_V=d$. The operation performed by the attention block is as follows $$Attention(Q,K,V) = softmax \left(\frac{QK^T}{d_K}\right)V$$ This gives us an output $U \in \mathbb{R}^{t' \times d}$, which is summed along the time axis and pass through fully connected feedforward layers. We use cross entropy loss to train this model. This architecture is shown in Figure \ref{fig:archs}. 
\subsubsection{Receptive Field of Recurrent Layer}
The motivation behind developing the Tiny-CRNN architecture for wakeword detection is to combine the scale-invariant features learned by convolutional layers with the long-term feature representations learned by the recurrent layers. This combination of temporal features with those learned via the convolutional layers is what gives the architecture its name, and as the following sections show, also greatly improve performance compared purely convolutional networks, which are in common use for wakeoword detection today. It is key, therefore, that the recurrent layers are given an input through which it possible to learn \emph{transitions} between different parts of the wakeword. We refer to our architecture in Figure \ref{fig:archs}, where the convolutional layers' outputs are of the size 10$\times$512. This means that 10 time steps are to be passed through the recurrent layers (in this case, GRU layers \cite{chung2014empirical}). The receptive field of the output of the convolutional front end, which corresponds to the time steps going into the recurrent layer, must contain a sufficient amount of context of a part of the wakeword, while simultaneously not having too much. If the receptive field of the time steps of the recurrent layer is too large, we are showing the recurrent layers the entire wakeword at each time step, thereby not allowing for any transitions between different parts (or \emph{phonemes}). Similarly, if the receptive field is too small compared to a non-trivial part of the wakeword, the recurrent layers would be rendered useless because they would be unable to learn the context around different parts of the wakeword correctly. Therefore, it is essential that the time steps capture relevant parts of the input to add any functionality to a straightforward CNN architecture. We find that a receptive field of around 30 input frames works best. The architecture in Figure \ref{fig:archs} has times steps with a receptive field of 28 input frames. 
\section{ Experimental Setup and Results}
\subsection{Training strategy}
We use internal research benchmark data for training and evaluation. 5388 hours of de-identified audio from a variety of devices in different acoustic conditions is used to to train our architectures, with a 50-50 split between human annotated positive (wakeword is present) and negative classes. We use a learning rate of $10^{-3}$ with the Adam optimizer, and train our models for 200k steps with a batch size of 2000 using Tensorflow \cite{tensorflow2015-whitepaper}. and train with cross-entropy loss. All models are trained with 0.3 dropout and batch normalization with either 64-bin (250k parameter budget) or 20-bin (50k parameter budget) LFBE features. We measure performance primarily by looking at the number of False Accepts (FAs) at a fixed Miss Rate (MR) of 15$\%$ as well as by plotting DET curves. These curves are plotted by calculating the False Detection Rate and MR at different thresholds, and then forming a curve with these two metrics as the axes.  Our evaluation dataset contains 0.9M positive and 1.1M  negative examples from across all device types. These devices include near field and far field ones, and vary broadly in terms of the quality of the audio front end (AFE). \comment{We explore two parameter budgets, 250k and 50k, comparing with baseline neural network architectures (CNNs and DNNs).}
\subsection{Does Attention Make a Difference?}
To show that our augmentation of a vanilla CRNN architecture with scaled dot product attention improves performance, we perform a comparison between two CRNNs with and without the attention mechanism, identical in all other respects. The results of this experiment are conclusive; we see a 5.2\% reduction in False Accepts across our test set. We hence set up all our CRNN models to include an attention block as described in Section \ref{crnn_models}. We cannot attribute this gain in performance to the small number of additional parameters that attention adds, since this is less than 1\% of the original parameter size. This shows us that attention provides clear and measurable improvements in reducing false accepts, and can be used as a building block to improve vanilla CRNN models. We believe there is value in augmenting CRNNs further with attention, and intend to explore mutli-headed and other types of attention mechanisms for CRNNs in future work.
\subsection{250k Parameter Budget Tiny-CRNNs}
We develop two CRNN architectures within this category, and both are trained with 64-bin LFBE features. The CRNN-239k (Figure \ref{fig:archs}) architecture uses 3 convolutional layers, and the receptive field of the output of the convolutional front end is 28. This outperforms our CNN-263k baseline by 25\% in terms of False Accepts with a 10\% decrease in parameter size. However, in terms of multiplies, the costs of the CRNN-239k architecture are nearly twice that of the CNN-263k architecture. We see that the multiplies are concentrated in the GRU layer, and therefore add another convolutional layer to downsample the input to reduce the multiplies contributed by the GRU layer. This architecture is called the CRNN-183k, and the improvement over our baseline CNN-263k reduces to 10\%, but the number of multiplies is comparable to CNN-263k.\begin{table}[!htpb]
\scalebox{0.8}{\begin{tabular}{
>{\columncolor[HTML]{FFFFFF}}c 
>{\columncolor[HTML]{FFFFFF}}c 
>{\columncolor[HTML]{FFFFFF}}c 
>{\columncolor[HTML]{FFFFFF}}c 
>{\columncolor[HTML]{FFFFFF}}c }
\toprule
{\color[HTML]{333333} \textbf{Model Name}}   & {\color[HTML]{333333} \textbf{FA Improvement}} & {\color[HTML]{333333} \textbf{Parameters}} & {\color[HTML]{333333} \textbf{Multiplies}} \\
\toprule
{\color[HTML]{333333} CNN-263k}                                      & {\color[HTML]{333333} Baseline}                & {\color[HTML]{333333} 263k}                & {\color[HTML]{333333} 5.25M}               \\
{\color[HTML]{333333} CNN-197k}                            & {\color[HTML]{333333}  -12\%}                & {\color[HTML]{333333} 197k}                & {\color[HTML]{333333} 3.5M}               \\
{\color[HTML]{333333} CNN-2.2M}                                      & {\color[HTML]{333333} 45\%}                    & {\color[HTML]{333333} 2.2M}                & {\color[HTML]{333333} 101M}               \\
{\color[HTML]{333333} \textbf{CRNN-239k}}           & {\color[HTML]{333333} \textbf{25\%}}                    & {\color[HTML]{333333} \textbf{239k}}                & {\color[HTML]{333333} \textbf{10.25M}}              \\
{\color[HTML]{333333} CRNN-183k}                                   & {\color[HTML]{333333} 10\%}                    & {\color[HTML]{333333} 183k}                & {\color[HTML]{333333} 5.73M}               \\
{\color[HTML]{333333} Delta-LFBE-CRNN-239k}                               & {\color[HTML]{333333} 12\%}                    & {\color[HTML]{333333} 239k}                & {\color[HTML]{333333} 10.2M}               \\
{\color[HTML]{333333} CRNN (Arik et. al.)}                                  & {\color[HTML]{333333} -21\%}                    & {\color[HTML]{333333} 106k (229k)}                & {\color[HTML]{333333} 2.2M}
\end{tabular}}
\end{table}
\begin{figure}[!htpb]
\centering
	\includegraphics[width=8cm,height=6cm]{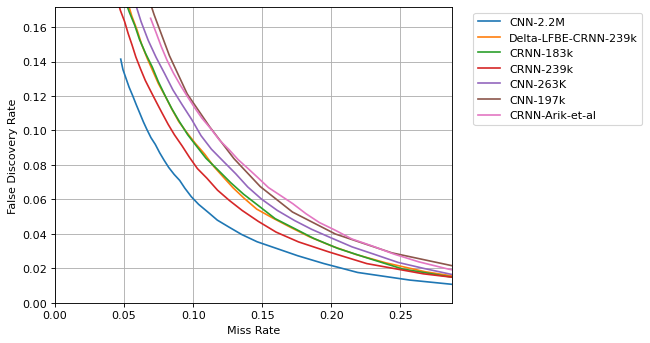}
\caption{\emph{250k Parameter Budget Results}}
\end{figure}
We also see that the best CRNN architecture within this parameter budget from \cite{DBLP:journals/corr/ArikKCHGFPC17}  is 21\% below the performance of even our baseline CNN. We would like to clarify that, based on our implementation, which involves folding the batchnorm layers into the convolution layers, the parameter size of this baseline CRNN is 106k, whereas the authors report it as 229k. Another source of variance is the fact that we used a $100 \times 64$ input as opposed to $40 \times 151$ input used by this paper. We consider it fair to place in this budget class of 250k, which is what the original implementation does. We also compare to a very large best-in-class CNN model  (CNN-2.2M) meant for full-power scenarios where small footprint is not a requirement, and see that it only performs 25\% better than our best CRNN model, while being nearly 10 times as expensive in terms of both parameters and multiplies.
\\ Recent work \cite{yixinpaper} has shown that using Delta-LFBE features (contiguously subtracting one input frame from the next) while building wakeword detection models helps make models invariant to Audio Front End (AFE) gain changes caused by different AFE algorithms on varied types of platforms. To implement this transformation, we increase the input frames to 101,  add a non-trainable convolutional layer with a $2\times1\times1$ kernel with the [$-1,1$] weights on top of the original convolutional layers, which gives us 100 frames, and we maintain the rest of the CRNN-239k model. Our attempt to induce robustness to AFE gain changes through these Delta-LFBE features (Delta-LFBE-CRNN-239k) is at the cost of performance on our test set, with only 12\% gain over the CNN-263k baseline as compared to our original non-delta LFBE-based CRNN-239k's 25\%. We leave confirmation of robustness to AFE gain changes in CRNNs through such data preprocessing to future work.
\subsection{50k Parameter Budget CRNNs}
\begin{table}[!htpb]
\scalebox{0.8}{\begin{tabular}{
>{\columncolor[HTML]{FFFFFF}}c 
>{\columncolor[HTML]{FFFFFF}}c 
>{\columncolor[HTML]{FFFFFF}}c 
>{\columncolor[HTML]{FFFFFF}}c 
>{\columncolor[HTML]{FFFFFF}}c }
\toprule
{\color[HTML]{000000} \textbf{Model Name}} & {\color[HTML]{000000} \textbf{\% FA Improvement}} & {\color[HTML]{000000} \textbf{Parameters}} & {\color[HTML]{000000} \textbf{Multiplies}} \\
\toprule
{\color[HTML]{000000} DNN-233k}                                               & {\color[HTML]{000000} Baseline}                & {\color[HTML]{000000} 233k}                & {\color[HTML]{000000} 233k}                \\
{\color[HTML]{000000} DNN-51k}                                                & {\color[HTML]{000000} -56\%}                   & {\color[HTML]{000000} 51k}                 & {\color[HTML]{000000} 51k}                 \\
{\color[HTML]{000000} CNN-28k}                                              & {\color[HTML]{000000} 24\%}                    & {\color[HTML]{000000} 28k}                 & {\color[HTML]{000000} 2.92M}               \\
{\color[HTML]{000000} \textbf{CRNN-89k}}                            & {\color[HTML]{000000} \textbf{40\%}}           & {\color[HTML]{000000} \textbf{89k}}        & {\color[HTML]{000000} \textbf{1.77M}}      \\
{\color[HTML]{000000} CRNN-58k}                                           & {\color[HTML]{000000} 32\%}                    & {\color[HTML]{000000} 58k}                 & {\color[HTML]{000000} 1.47M}           \\   
\comment{{\color[HTML]{000000} Delta-LFBE-CRNN-89k}                                            & {\color[HTML]{000000} 26\%}                    & {\color[HTML]{000000} 89k}                 & {\color[HTML]{000000} 1.77M} }       
\end{tabular}}
\end{table}
\begin{figure}[!htpb]
\centering
	\includegraphics[width=8cm,height=6cm]{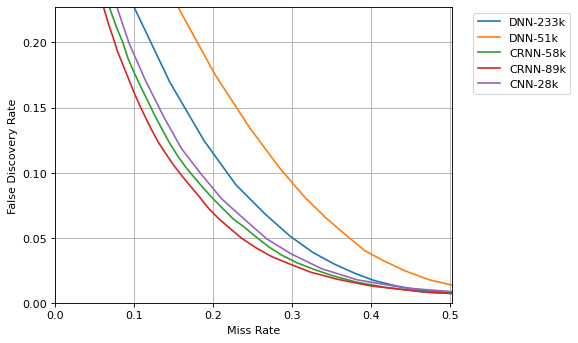}
\caption{\emph{50k Parameter Budget Results}}
\end{figure}

For environments that are even more compute-constrained, such as TV remotes and smartwatches, we develop a set of CRNN architectures with an even smaller footprint. We perform this reduction in model size by switching to 20-bin LFBE features instead of 64-bin features and reducing the number of filters in the convolutional layers. We compare this to low parameter DNN and CNN models. We see that, even over a 233k parameter DNN model (DNN-233k) (the reason we compare to this model is that the multiplies of our model are much higher, even though the parameter size is lower, which makes for a reasonable comparison), our CRNN-58k  performs 32\% better in terms of FAs, and even better when compared to DNN-51k. We also show that we comfortably beat a smaller CNN model (CNN-28k), and even though this might be expected considering that our model has twice as many parameters, this is not an obvious result if we notice that the CNN-28k has nearly twice the number of multiplies when compared to our model. We also develop a slightly larger model, the CRNN-89k, where the advantages over DNNs and CNNs become even more pronounced. We see, however, that the multiplies in a DNN are linear in the number of parameters, while this is strictly not true for CNN and RNN layers. We therefore see a six-fold increase in multiplies when compared to standard DNN models with our CRNN-58k.

\section{Endpoint Error and Latency} Accurate prediction of wakeword endpoints is an important task for the wakeword detection model \cite{Jose2020AccurateDO}. This is especially important in the case of voice assistant systems, which would need to accurately estimate the wakeword start and end to be able to send audio to downstream services for further processing to actually perform whatever task is actually required after keyword spotting takes place. To estimate this, we measure the delta between start and end indices of our CRNN-239k and CNN-263k models with respect to a 2-stage DNN-HMM baseline \cite{Jose2020AccurateDO}. The deviations with respect to a 2-stage model in start indices are 204ms and 212ms and end indices are 204ms and 223ms for CRNN-239k and CNN-263k respectively, a slight improvement shown over the CNN.  We use the fact that our 2-stage DNN-HMM baseline has a mean 50ms latency, which we add to the mean delta in end indices for both our models to get their mean latency, which is 218ms and 172ms for CRNN-239k and CNN-263k. This increased latency in CRNNs, we hypothesize, is due to the overhead of recurrent layers. However, this difference in latency of approximately 50ms would not cause any appreciable difference in real-world usage when used in a real-time edge computing system where the wakeword under consideration,``Alexa", has a median length of 700ms. The computation of this latency is tentative, often, with optimizations built into various platforms that models are able to leverage, we can reduce the latency further.  \\
\section{Streaming Inference in CRNNs}
\subsection{Resetting States for CRNN Recurrent Layers}
In previous work about keyword spotting, and the results above, we always consider test-time inference to be on fixed-length data, with the same number of frames as the model was originally trained on.  However, when doing wakeword detection in the wild, the device our model runs on is always on, and inference must be done on-device with our trained models on streaming audio. Therefore, a continuous stream of audio passes through the devices. This is a different case than of carefully prepared training examples which can be fed in batches to the convolutional front end, and requires careful deliberation. In the case of convolutional layers, an issue of efficiency arises, since convolutions over different overlapping parts of the stream would needlessly re-compute expensive operations. This can be mitigated through the use of a ring buffer, as described in \cite{Rybakov2020StreamingKS}, which enables efficient streaming convolutions by saving computations. However, in the case of recurrent layers (GRUs) present in our model, an issue of correctness arises along with that of efficiency.  A network trained with $t$ input frames has $t'$ time steps (as described in Section \ref{crnn_models}) as output from the convolutional layer to be passed into the recurrent layer, with a new one generated every (say) $k$ frames. However, our trained recurrent layers take only $t'$ time steps. If we were to feed the recurrent network $t'$ non-overlapping time steps, we would miss detections, and if we were to use overlapping blocks of $t'$ time steps at a stride of $1$, we would need to reset states of the recurrent layer every time a new time step is output by the convolutional layers.  \\
\subsubsection{Parallel decoders for Streaming Inference}
\begin{figure}[!htb]
\centering
	\includegraphics[width=8cm,height=14cm]{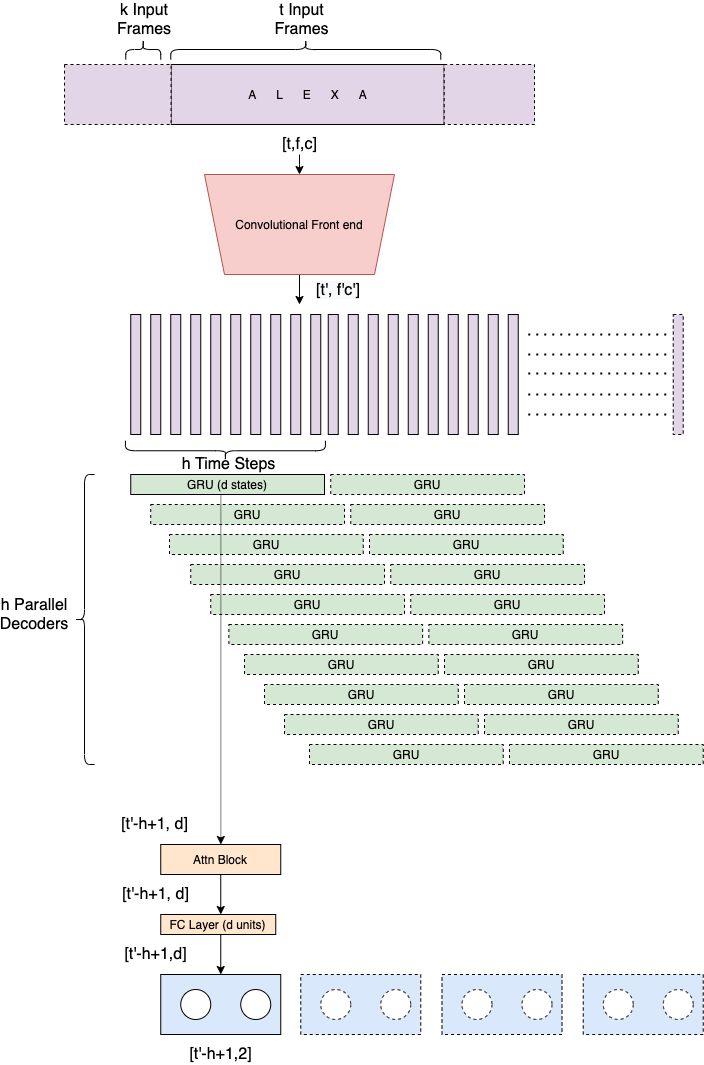}
\caption{\emph{Parallel Decoders for Streaming Inference}}
\label{fig:inference}
\end{figure}

To avoid both the issues, we use parallel decoders. We can use $t'$ GRU decoders in a semi-parallelized fashion, resetting only after $t'$ sets of $t'$ time steps pass through them. Say $t'=10$, this would mean having the first GRU take in steps $1$-$10$, but instead of resetting the first GRU's states, we would have another identical GRU decoder which takes in steps $2$-$11$, and another $8$ GRUs, which take in all such combinations until steps $10$-$19$. Only at this point do we reset all the GRUs' states in parallel, and use them starting with inputs $11$-$20$ and so on. We illustrate this in Figure \ref{fig:inference}. This method enables us to both correctly identify detections, as well as have low latency outputs since a new posterior is created after each new time step arrives at the recurrent layer. However, this is only a realistic scenario with models that have a limited number of time steps going into the recurrent layers. If the model is such that there are a large number of time steps passing through the recurrent layer, the number of GRU decoders would scale linearly, which would take up a significant amount of memory.\\

\subsubsection{A Higher Dimensional GRU for Streaming Inference}
We can also show that by making use of vectorization, which may be available to us on even low footprint device platforms, it is possible to implement a GRU (or any other recurrent network) that can process multiple inputs at the same time. This is similar to how batch inference is done in frameworks such as Tensorflow. We first wait for $2t'-1$ time steps and convert them into overlapping inputs, each of size $t'$. To make use of this however, it is important to note that we can compute overlapping input matrices very cheaply. It is straightforward, once we have $2t;-1$ time steps of dimensions $f'c'$, to convert them into a matrix $X_t$ of shape $t' \times t' \times f'c'$, with overlaps corresponding to the inputs to the parallel decoder described above. We now look at one of the operations that take place in a GRU, that conducted by the update gate. 
$$z = \sigma(W_z . X_t + U_z . H_{t-1} + b_z)$$ In the normal non-streaming case, for a GRU with $d$-dimensional hidden states, the trained parameter matrices $W_z$ and $U_z$ are of the shapes $d \times t'$ and $d \times d$, while the hidden states are vectors of dimension $d \times 1$.  We can expand this to $t' \times d \times 1$  which allows us to have, instead of $d$, $t' \times d$ hidden states built into the GRU. It is important to remember that this type of GRU is only used during inference, so we always know the value of $t'$ from the number of time steps going into a regular GRU while training. Now, when the operations are performed, vectorization leads to an output of $t' \times d \times 1$ instead of $d \times 1$. Correspondingly the entire GRU outputs a matrix of shape $t' \times d \times 1$, whereupon passing the output through the rest of the network gives us $t'$ posteriors from $2t'-1$ time steps output by the convolutional front end, which is the same as in the previous section with parallel decoders. In this method, however, we are able to use the same parameter matrices throughout, instead of having $t'$ copies.  
\\ Overall, both these methods have their drawbacks. The former requires multiple copies of the GRU, increasing the space taken on device, but inference through the latter part of the network is not a bottleneck. In the latter approach, however, while inference through the recurrent layers can happen in parallel, the rest of the neural network is a bottleneck in terms of speed. This makes evaluating the speed and computation tradeoffs of these models a complex endeavor. We leave the implementation and profiling of the methods discussed in this section to future work.

\section{Conclusion}
In this paper, we describe and show the efficacy of the Tiny-CRNN (Tiny Convolutional Recurrent Neural Networks) architecture for small footprint wakeword detection over established CNN (by 25\% decrease in FAs at a 250k parameter budget) and DNN architectures (by 32\% decrease in FAs within a 50k parameter budget). We show that simple scaled dot product attention improves CRNN performance while adding negligible parameters. We investigate strategies for efficient streaming inference and robustness to differences in audio front end gain. Through latency comparisions, we show that models based on the Tiny-CRNN architecture are production ready in similar environments as the ones in which CNN models are deployed today, and overall, establish that our approach is both computationally efficient and highly performant. We believe that this combination of two extremely popular architectures (CNN and RNNs) is an effective tool for keyword spotting, and may perhaps be used in future audio-focused machine learning work.

\bibliographystyle{IEEEbib}
\bibliography{strings,refs}

\end{document}